\documentclass[letterpaper, 10 pt, conference]{ieeeconf}  
\IEEEoverridecommandlockouts                              
\title{\LARGE \bf Information-Guided Safe Reinforcement Learning\\
for Autonomous Gas Source Localization using sUAS} 

\author{Sachin Giri$^{1}$, Thomas Zhao$^{2}$, Matthew Huynh$^{3}$ and YangQuan Chen$^{4,*}$
\thanks{The authors were supported by the Center for Methane Emission Research and Innovation (CMERI) through the Climate Action Seed Funds grant at University of California, Merced (2023-2026).}
\thanks{$^{1}$Electrical Engineering \& Computer Science Graduate Program, University of California, Merced,
        5200 N. Lake Rd, Merced, CA 95343, USA
        \faEnvelope~{\tt\small sachingiri@ucmerced.edu}}%
\thanks{$^{2}$Dept. of Mechanical Engineering, University of California, Merced,
        5200 N. Lake Rd, Merced, CA 95343, USA
        \faEnvelope~{\tt\small tzhao26@ucmerced.edu}}%
\thanks{$^{3}$Dept. of Mechanical Engineering, University of California, Merced,
        5200 N. Lake Rd, Merced, CA 95343, USA
        \faEnvelope~{\tt\small mhuynh35@ucmerced.edu}}%
\thanks{$^{4}$Dept. of Mechanical Engineering, University of California, Merced,
        5200 N. Lake Rd, Merced, CA 95343, USA
        \faEnvelope~{\tt\small ychen53@ucmerced.edu}}%
}

\usepackage{graphicx}
\usepackage{amsmath}
\usepackage{placeins}
\usepackage{hyperref}
\usepackage{amssymb}  
\usepackage{float}  
\usepackage{fancybox}
\usepackage{tcolorbox}
\usepackage{tikz}
\usetikzlibrary{tikzmark}
\usepackage[utf8]{inputenc}
\usepackage[linesnumbered,ruled,vlined]{algorithm2e}
\usepackage{color}
\usepackage{fontawesome}  
\usepackage{subfigure}
\usepackage{cite}

\newtheorem{remark}{Remark}

\begin{document}

\maketitle


\begin{abstract}
The autonomous localization of fugitive gas emissions using small Unmanned Aircraft Systems (sUAS) constitutes a fundamentally ill-posed inverse problem. In turbulent atmospheric boundary layers, highly intermittent scalar concentration fields violate the assumptions of classical gradient-based navigation, causing data-driven estimators to suffer from severe noise and spurious local minima. To address these challenges, we introduce an Information-Guided Safe Reinforcement Learning framework evaluated within a custom, GPU-accelerated 3D simulation environment coupling an Eulerian wind solver with a Lagrangian puff dispersion model. We identify a critical vulnerability in deterministic information-seeking planners-a \textit{Gramian bias} where agents act greedily upon flawed early estimates, starving the estimator of spatial diversity. To systematically break this degeneracy, our architecture integrates a classical empirical observability Gramian (EMGR) planner with a learned Soft Actor-Critic (SAC) exploratory policy. A deterministic meta-supervisor actively monitors estimator reliability via Kullback-Leibler (KL) divergence, dynamically blending deterministic exploitation with learned exploration to steer the sUAS into high-information zones. Trained via a progressive curriculum and safeguarded by a strictly enforced Robust Control Barrier Function (RCBF), our RL framework achieves nearly 80\% localization success on complex, mobile sources—drastically outperforming classical baselines ($\sim$30\%)—while ensuring zero safety violations.
\end{abstract}

\addvspace{0.5\baselineskip}
\section{INTRODUCTION}

Source seeking utilizing mobile sensor platforms is critical for environmental monitoring, industrial leak detection, and disaster response. The overarching objective is to navigate an autonomous agent to identify the origin of a diffusing substance using sparse, sequential measurements. In realistic outdoor conditions, turbulent mixing dominates molecular diffusion, creating a highly intermittent scalar field. This intermittency renders classical gradient-ascent methods (chemotaxis) \cite{keller1971model} highly susceptible to local maxima. To overcome these limitations, strategies such as anemotaxis (upwind search) \cite{prabowo2023integration}, infotaxis (entropy reduction) \cite{vergassola2007infotaxis}, entrotaxis \cite{hutchinson2018entrotaxis}, and fluxotaxis \cite{spears2012physicomimetics} were developed. 

However, even advanced probabilistic approaches struggle to precisely estimate the source parameters. Estimating the source location $\theta \in \mathbb{R}^3$ via data-driven methods like Maximum Likelihood Estimation (MLE) is inherently an ill-posed inverse problem \cite{golan2017foundations}. Sensor noise, ambient bias, and extreme signal sparsity conspire to violate Hadamard's conditions for well-posedness, yielding highly inaccurate early estimates. Prior source-seeking formulations have heavily utilized metrics like the Cramér-Rao Lower Bound (CRLB) \cite{Nehorai1995, porat1996} and the Fisher Information Matrix (FIM) \cite{ucinski2004optimal} to guide trajectory planning. 

We previously demonstrated \cite{giri2025optimal} that trajectory planning guided by the empirical observability Gramian (EMGR) \cite{himpe2018emgr, powel2020empiricalobservabilitygramianstochastic} provides superior convergence over CRLB and FIM by directly quantifying the observability of the nonlinear system \cite{hollenbeck2023digital, krener2009measures}. However, a critical vulnerability remains: deterministic planners computing trajectories based solely on current parameter estimates are prone to a secondary degeneracy. Acting greedily upon a flawed MLE draws the agent into a stalled state, starving the estimator of the spatial diversity required to correct itself.

Recent literature increasingly uses Reinforcement Learning (RL) and domain knowledge to navigate complex plumes \cite{wiedemann2021robotic, rando2025q}. Active inference frameworks for drones have also demonstrated that active sampling drastically reduces parameter uncertainty \cite{van2026actively}. We posit that sequential maximization of information along a trajectory can actively regularize this ill-posed inverse problem if the agent balances deterministic exploitation with spatial exploration. 

To overcome these limitations, we propose a Residual Safe RL architecture enveloping classical EMGR methods within an RL regularizer governed by a deterministic Meta-Supervisor. By continuously monitoring the Kullback-Leibler (KL) divergence \cite{kullback1951information} between empirical observations and estimator predictions, this supervisor dynamically blends deterministic EMGR velocities with learned exploratory actions from an off-policy Meta-Soft Actor-Critic (Meta-SAC) policy \cite{haarnoja2018soft, finn2017model, rakelly2019efficient}. Actively navigating the sUAS toward high-information zones systematically breaks the MLE degeneracy and reduces estimator variance. For real-world deployment safety, we integrate Safe RL via Control Barrier Functions (CBFs) \cite{ames2019control, hsu2015control, cheng2019end, dalal2018safe, emam2022safe}. Advancing beyond data-driven CBF approximations \cite{folkestad2020data, manzoor2023vehicular}, we formulate a kinematic Robust Control Barrier Function (RCBF) \cite{xu2015robustness, emam2019robust}. This safety layer solves a Quadratic Program (QP) using a highly optimized operator splitting solver \cite{stellato2020osqp} at each timestep, strictly mapping the blended policy into a provably safe action space while accounting for worst-case disturbances.

A major supporting contribution is a highly optimized, GPU-accelerated 3D simulation environment built on the NVIDIA Warp framework \cite{nvidia_warp}. By unifying computer graphics fluid simulations \cite{foster1996realistic, stam2023stable, bridson2015fluid} with near-field atmospheric puff models \cite{jia2025fast}, we provide a mathematically rigorous testbed for training turbulent source-seeking policies.

The remainder of this paper is structured as follows. Section \ref{environment} details the 3D simulation environment, and Section \ref{estimation} formulates the ill-posed inverse problem and EMGR. Section \ref{Rl-framework} presents the Information-Guided Safe RL architecture and RCBF. Finally, Section \ref{sec:results} discusses simulation results, followed by concluding remarks in Section \ref{conclusion}.

\section{GPU-ACCELERATED 3D SIMULATION ENVIRONMENT} \label{environment}

\begin{figure}[htbp]
    \centering
    \includegraphics[width=0.75\linewidth]{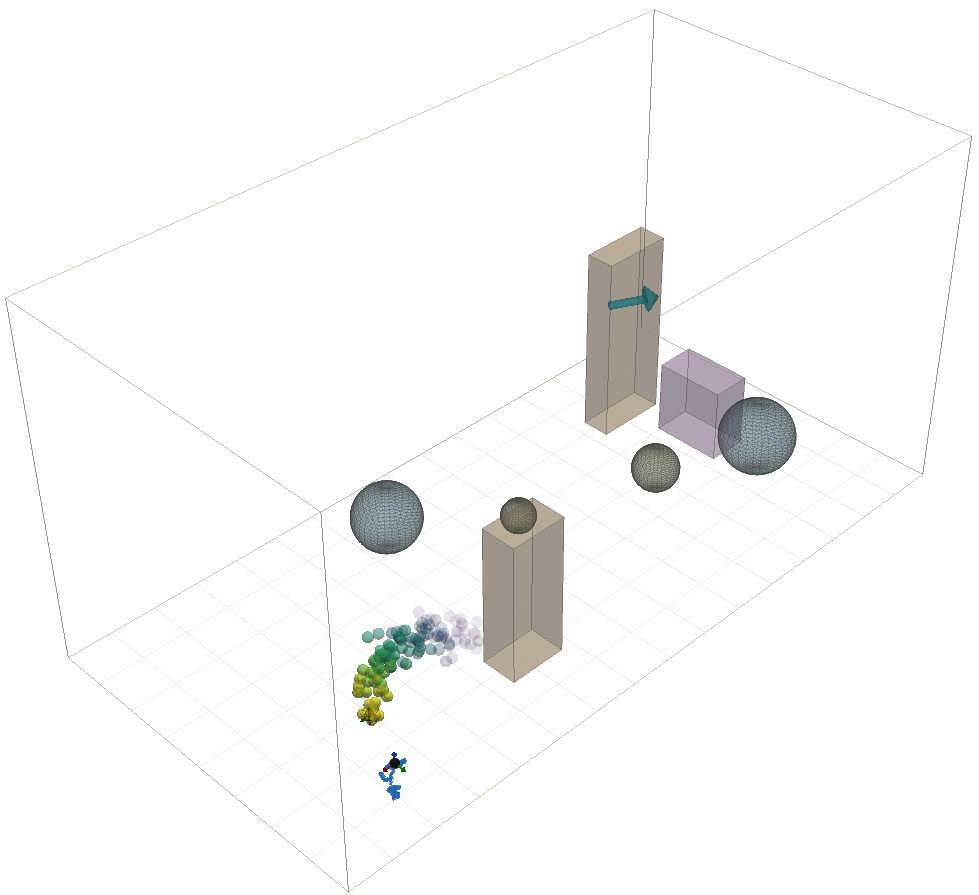}
    \caption{3D simulation environment built with NVIDIA Warp, demonstrating the Lagrangian puff dispersion model (color-coded by concentration) navigating around physical obstacles.}
    \label{fig:3d_env}
\end{figure}

To train and evaluate the proposed framework under realistic atmospheric intermittency, we developed a custom simulation environment coupling an Eulerian wind solver with a Lagrangian puff dispersion model. This hybrid approach captures both macroscopic wind dynamics and crucial sub-grid scalar intermittency, leveraging NVIDIA Warp \cite{nvidia_warp} for high-performance GPU execution (Fig.~\ref{fig:3d_env}).

\subsection{Eulerian Stable Fluids Solver}
To simulate the wind environment, the domain $\Omega = [0, L_x] \times [0, L_y] \times [0, L_z]$ is discretized into a collocated grid. The background wind velocity vector field, denoted as $\mathbf{u}(\mathbf{x},t) \in \mathbb{R}^3$, is governed by the incompressible Navier-Stokes equations \cite{foster1996realistic}:
\begin{align}
    \frac{\partial \mathbf{u}}{\partial t} + (\mathbf{u} \cdot \nabla)\mathbf{u} &= -\nabla p + \nu\nabla^2\mathbf{u} + \mathbf{f}, \label{eq:momentum} \\
    \nabla \cdot \mathbf{u} &= 0, \label{eq:incompressibility}
\end{align}
where $\frac{\partial \mathbf{u}}{\partial t}$ represents the local acceleration, $(\mathbf{u} \cdot \nabla)\mathbf{u}$ captures the advection of momentum, $-\nabla p$ denotes the kinematic pressure gradient, $\nu\nabla^2\mathbf{u}$ accounts for viscous diffusion, and $\mathbf{f}$ encapsulates external forcing.

To achieve unconditional numerical stability at real-time rendering frame rates, we employ the Stable Fluids methodology \cite{Stam1999, stam2023stable, bridson2015fluid} utilizing operator splitting. The advection step computes an intermediate velocity field $\mathbf{u}^*$ via semi-Lagrangian backward particle tracing. To enforce the incompressibility constraint \eqref{eq:incompressibility}, we apply a Helmholtz-Hodge projection \cite{chorin1968numerical}, which decomposes the intermediate field into a divergence-free velocity field and a corrective pressure gradient. The resulting discrete Poisson equation is solved efficiently on the GPU using Red-Black Successive Over-Relaxation (RB-SOR) \cite{Young1971}. External forcing $\mathbf{f}$ is layered using an Ornstein-Uhlenbeck (OU) stochastic process \cite{uhlenbeck1930theory} to mathematically mimic realistic gusty atmospheric boundary layer behavior and turbulence spectra \cite{Stull1988}. The wind shear profile follows a power-law with exponent $\alpha_s = 0.25$ for neutral class D, and Langevin dispersion coefficients $\ell_\alpha$ are assigned per Pasquill--Gifford stability class \cite{NOAA_READY}.

Boundary conditions are strictly enforced to ensure physical realism. Inside embedded obstacles, velocity is explicitly zeroed ($\mathbf{u} = \mathbf{0}$). At the ground plane ($z=0$), a no-penetration condition prevents vertical flow ($w=0$), while surface friction within the atmospheric boundary layer is modeled via localized proportional drag in the lowest grid cells. During the projection step, solid interfaces dictate a homogeneous Neumann boundary condition ($\frac{\partial p}{\partial \mathbf{n}} = 0$) for the Poisson solver, preventing corrective pressure gradients from pushing mass through impermeable surfaces. Conversely, the open domain limits (lateral faces and sky) are treated as free-outflow boundaries with a Dirichlet condition ($p = 0$) \cite{bridson2015fluid, fedkiw2001visual}, ensuring wind and turbulent structures exit cleanly without artificial mass accumulation (Fig.~\ref{fig:wind_speed}).

\subsection{Lagrangian Gaussian Puff Model}
Using an Eulerian grid to directly track scalar gas concentrations introduces severe numerical dissipation, artificially smearing the sharp filaments and intermittent bursts characteristic of real-world leaks. To preserve this crucial sub-grid intermittency, we transition to a Lagrangian viewpoint \cite{jia2025fast}. We model the continuous leak at source location $\theta \in \mathbb{R}^3$ as a rapid sequence of discrete gas puffs. A new puff is released every $\tau_e$ seconds, carrying a fixed mass $m_i = Q \tau_e$, where $Q$ is the continuous emission rate.

Each puff's centroid $\mathbf{x}_i$ is updated using a hybrid two-part kinematic scheme. A second-order Runge-Kutta (RK2) predictor-corrector scheme \cite{press2007numerical} integrates the macroscopic Eulerian wind, while sub-grid turbulence is modeled via random Langevin noise \cite{Thomson1987}. The spatial position $\mathbf{x}_i$ of the $i$-th puff at time step $n+1$ is updated via:
\begin{align}
    \mathbf{k}_1 &= \mathbf{u}(\mathbf{x}_i^n) + a_b e^{-c_b a_i} \hat{\mathbf{z}} \label{eq:k1} \\
    \mathbf{k}_2 &= \mathbf{u}\left(\mathbf{x}_i^n + \frac{\Delta t}{2}\mathbf{k}_1\right) + a_b e^{-c_b a_i} \hat{\mathbf{z}} \label{eq:k2} \\
    \mathbf{x}_i^{n+1} &= \mathbf{x}_i^n + \Delta t \mathbf{k}_2 + \sqrt{\Delta t} \begin{bmatrix} \ell_x \eta_x \\ \ell_y \eta_y \\ \ell_z \eta_z \end{bmatrix} \label{eq:puff_update}.
\end{align}
Here, $\mathbf{u}(\cdot)$ is the localized Eulerian wind velocity. The $a_b$ term represents the initial buoyancy acceleration; for methane ($\rho_{\mathrm{CH_4}} = 0.657$\,kg/m$^3$) relative to air, $a_b \approx 4.55$\,m/s$^2$. As the puff ages ($a_i$), mixing diminishes this buoyant advantage, modeled by the empirical entrainment decay constant $c_b$. For the Langevin noise in \eqref{eq:puff_update}, we utilize independent random draws $\eta_\alpha \sim \mathcal{N}(0,1)$ scaled by intensity coefficients $\ell_\alpha$ derived from Pasquill-Gifford stability classes \cite{Gifford1961}.

The spatial concentration of each puff follows a 3D Gaussian distribution. As a puff ages, turbulence stretches it, causing its anisotropic variance to grow linearly: $\sigma_\alpha^2 = \sigma_0^2 + 2 K_{D,\alpha} a_i$, where $\sigma_0$ is the initial size and $K_{D,\alpha}$ is the turbulent eddy diffusivity. The total spatial concentration $C(\mathbf{r}, t)$ measured at any sensor position $\mathbf{r} = (x, y, z)$ is the overlapping sum of all active puffs. To enforce a physical no-flux boundary condition at the earth's surface ($z=0$), we employ the method of images \cite{stockie2011mathematics}, mathematically simulating a mirror-image ghost puff to conserve mass. The final concentration field is computed as:
\begin{equation}
\resizebox{0.91\linewidth}{!}{%
    $\displaystyle C(\mathbf{r}, t) = \sum_{i} \frac{m_i}{(2\pi)^{3/2} \sigma_x \sigma_y \sigma_z} \exp\left(-\frac{(x-x_i)^2}{2\sigma_x^2} - \frac{(y-y_i)^2}{2\sigma_y^2}\right) G_z$%
}
\label{eq:puff_conc_final}
\end{equation}
where the vertical reflection term is given by:
\begin{equation}
    G_z = \exp\left(-\frac{(z-z_i)^2}{2\sigma_z^2}\right) + \exp\left(-\frac{(z+z_i)^2}{2\sigma_z^2}\right).
\end{equation}

\begin{figure}[htbp]
    \centering
    \subfigure[Mean Eulerian wind speed $|\text{u}|$ (max 3.7 m/s) around obstacles.]{%
        \includegraphics[width=0.48\linewidth]{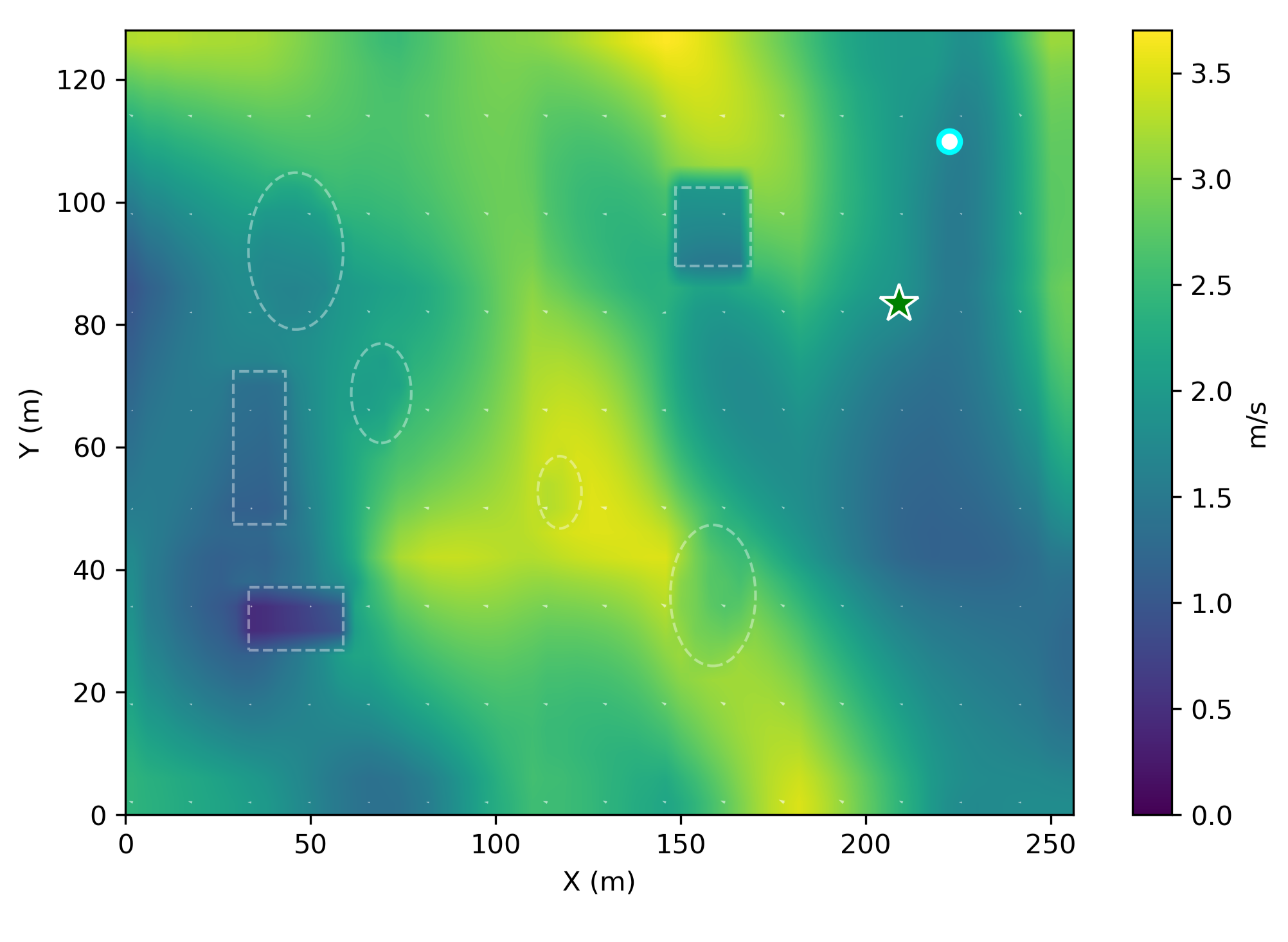}%
        \label{fig:wind_speed}%
    }\hfill 
    \subfigure[Column-integrated Lagrangian concentration C(r,t) at t=12s 
showing intermittent plume structure.]{%
        \includegraphics[width=0.48\linewidth]{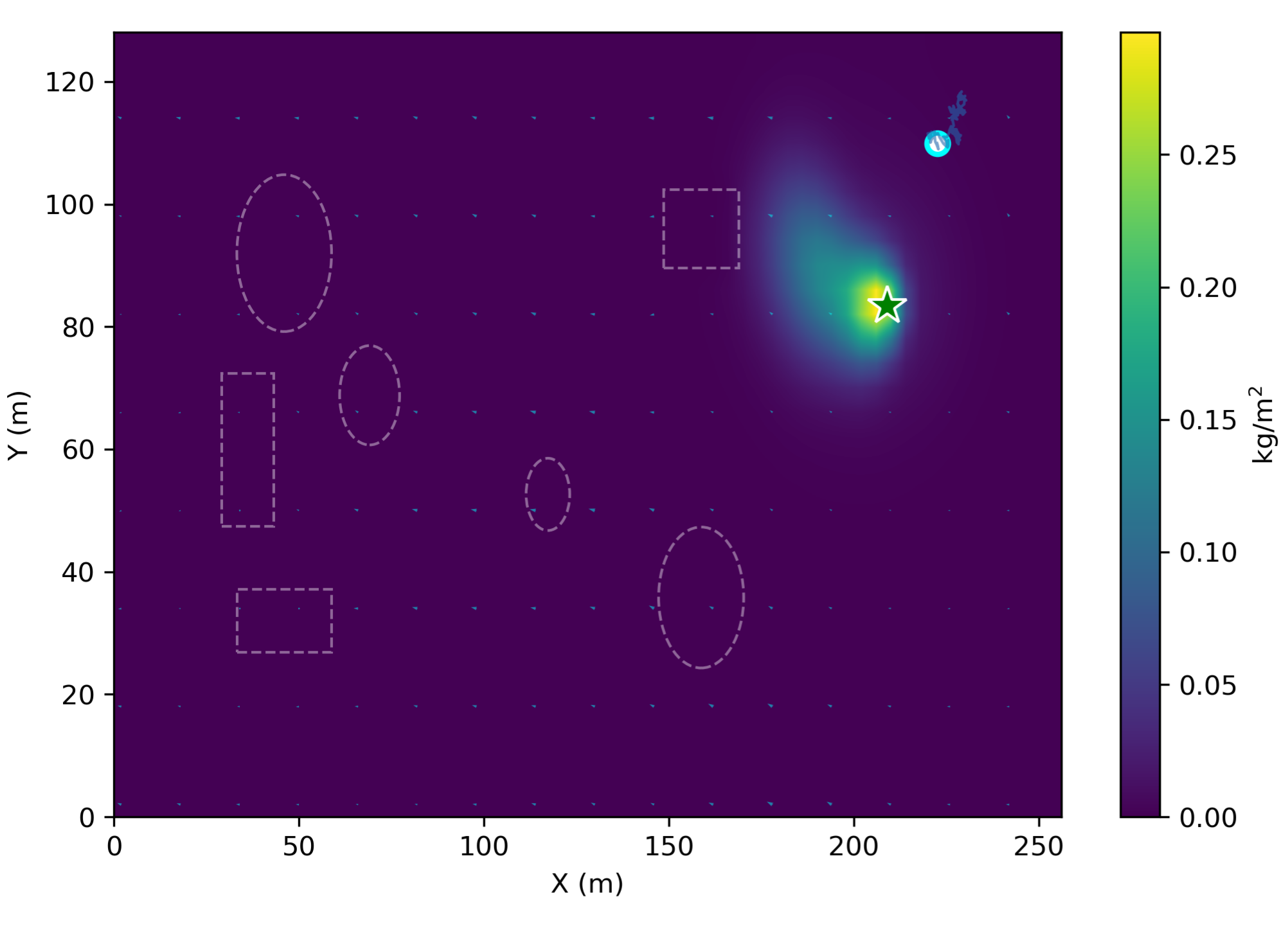}%
        \label{fig:concentration}%
    }
    \caption{2D cross-sections of the coupled simulation environment.}
    \label{fig:2d_maps}
\end{figure}

By mathematically decoupling the microscopic gas transport from the macroscopic grid, this formulation eliminates Eulerian numerical smearing. Preserving these sharp concentration gradients is critical for the RL agent to navigate realistically (Fig.~\ref{fig:concentration}), directly narrowing the reality gap for successful sim-to-real (sim2real) transfer to physical sUAS hardware.

\section{ILL-POSED ESTIMATION AND OBSERVABILITY} \label{estimation}

\subsection{The Ill-Posed Inverse Problem}

Let $\mathcal{P} \subset \mathbb{R}^3$ denote the parameter space of candidate source locations, $\theta \in \mathcal{P}$. Along a trajectory $\mathcal{R} = [\mathbf{r}_1, \ldots, \mathbf{r}_M]^T$, the sUAS collects scalar measurements $\mathcal{Y} = [y_1, \ldots, y_M]^T$:
\begin{equation}
    y_k = C(\mathbf{r}_k, t_k;\, \theta) + \varepsilon_k, \qquad \varepsilon_k \sim \mathcal{N}(0,\, \sigma_e^2),
    \label{eq:observation_model}
\end{equation}
with $C(\cdot)$ from \eqref{eq:puff_conc_final}. The nonlinear forward operator $\mathcal{F}: \mathcal{P} \to \mathbb{R}^M$ maps $[\mathcal{F}(\theta)]_k = C(\mathbf{r}_k, t_k;\, \theta)$. Recovering $\theta$ from $\mathcal{Y}$ is severely ill-posed \cite{golan2017foundations}: spatial sparsity of $C(\mathbf{r},t)$ renders distinct hypotheses indistinguishable, while small perturbations in $\varepsilon_k$ produce unbounded variations in $\hat{\theta}$. In 3D turbulent conditions, the concentration gradient $\nabla_{\mathbf{r}} C$ is dominated by advection and points predominantly \textit{downwind}, away from $\theta$, rather than toward it.

\subsubsection{Maximum Likelihood Estimation} Over a sliding window of $N$ measurements, let $\mathbf{Y} \in \mathbb{R}^N$ be the observation vector and $\mathbf{a}(\theta) \in \mathbb{R}^N$ the predicted concentrations, $a_j(\theta) = C(\mathbf{r}_j, t_j;\, \theta)$. The MLE maximizes the mean-subtracted projected correlation:
\begin{equation}
    \hat{\theta}_{\text{MLE}} = \arg\max_{\theta \in \mathcal{P}} \; \frac{\left(\mathbf{Y}^T \mathbf{a} - \frac{1}{N} \mathbf{1}^T \mathbf{Y} \cdot \mathbf{1}^T \mathbf{a}\right)^2}{\mathbf{a}^T \mathbf{a} - \frac{1}{N}(\mathbf{1}^T \mathbf{a})^2 + \varepsilon_{\text{mle}}},
    \label{eq:proj_mle}
\end{equation}
where $\mathbf{1} \in \mathbb{R}^N$ is the ones vector and $\varepsilon_{\text{mle}}$ regularizes the denominator. To capture long-range wind dependence in $\mathbf{a}(\theta)$, the wind history $\{\mathbf{u}_k\}$ is preprocessed with a Gr\"{u}nwald--Letnikov fractional filter \cite{scherer2011grunwald} of order $\alpha$:
\begin{equation}
    w_0^{(\alpha)} = 1, \quad w_j^{(\alpha)} = w_{j-1}^{(\alpha)} \!\left(1 - \frac{\alpha + 1}{j}\right), \;\; j \geq 1,
    \label{eq:gl_weights}
\end{equation}
yielding filtered wind $\tilde{\mathbf{u}}_k = (\bar{\mathbf{w}}^{(\alpha)} * \mathbf{u})_k$ with $\ell_1$-normalized weights $\bar{\mathbf{w}}^{(\alpha)}$.
Since puff advection integrates wind over its travel time, we use $\alpha < 0$ (fractional integration) to introduce long-memory smoothing. The order $\alpha \in [-0.9,\, 0]$ is selected via grid search over deterministic survey trajectories.

\subsubsection{Particle Filter for Hypothesis Diversity} The MLE provides a point estimate and cannot represent multimodal uncertainty. We complement this frequentist perspective with a Bayesian particle filter (PF) maintaining a weighted ensemble $\{\theta^{(i)}, w^{(i)}\}_{i=1}^{N_p}$ over $\mathcal{P}$. Upon measurement $y_k$, the importance weights are updated:
\begin{equation}
    w_k^{(i)} \propto w_{k-1}^{(i)} \cdot \frac{1}{\sqrt{2\pi}\,\sigma_e} \exp\!\left(-\frac{\bigl(y_k - C(\mathbf{r}_k;\, \theta^{(i)})\bigr)^2}{2\sigma_e^2}\right),
    \label{eq:pf_update}
\end{equation}
followed by normalization. Systematic resampling with jitter is triggered when the effective sample size (ESS), defined as $\mathrm{ESS}_k = 1 / \sum_i (w_k^{(i)})^2$ \cite{arulampalam2002tutorial}, falls below a threshold. The ESS quantifies weight degeneracy: when few particles carry most of the mass, $\mathrm{ESS}_k \ll N_p$, indicating that the filter's approximation has collapsed. The PF estimate is $\hat{\theta}_{\text{PF}} = \sum_i w_k^{(i)} \theta^{(i)}$, and its entropy $H_k = -\sum_i w_k^{(i)} \ln w_k^{(i)}$ signals convergence. By leveraging both the frequentist MLE and Bayesian PF---competing yet complementary paradigms---we hedge against each method's weaknesses.

\subsubsection{KL Divergence Model-Mismatch Detector} To monitor $\hat{\theta}_{\text{MLE}}$ reliability online, we compare predicted and observed concentrations over a buffer of $N_b$ measurements via the KL divergence \cite{kullback1951information}. The min-max normalized observed measurements $\mathbf{c}_{\text{obs}} \in \mathbb{R}^{N_b}$ from \eqref{eq:observation_model} and predicted concentrations $\mathbf{c}_{\text{pred}} \in \mathbb{R}^{N_b}$ evaluated at $\hat{\theta}_{\text{MLE}}$ via \eqref{eq:puff_conc_final} are converted to distributions through softmax with temperature $\tau$:
\begin{equation}
    q_j = \frac{e^{c_{\text{obs},j} / \tau}}{\sum_{l} e^{c_{\text{obs},l} / \tau}}, \quad p_j = \frac{e^{c_{\text{pred},j} / \tau}}{\sum_{l} e^{c_{\text{pred},l} / \tau}},
    \label{eq:softmax_kl}
\end{equation}
and the mismatch signal is:
\begin{equation}
    D_{\text{KL}}(q \,\|\, p) = \sum_{j=1}^{N_b} q_j \ln \frac{q_j}{p_j}.
    \label{eq:kl_mismatch}
\end{equation}

Large $D_{\text{KL}}$ indicates an unreliable $\hat{\theta}_{\text{MLE}}$ and triggers the exploration mechanism in Section~\ref{Rl-framework}. Regardless of the estimator, the quality of $\hat{\theta}$ is bounded by the informational content of $\mathcal{R}$. This motivates the observability analysis next.

\subsection{Empirical Observability Gramian (EMGR)}

To actively regularize the inverse problem, the trajectory $\mathcal{R}$ must maximize the sensitivity of $\mathcal{F}(\theta)$ to perturbations in the parameter space. In our previous work \cite{giri2025optimal}, we demonstrated that the empirical observability Gramian \cite{lall1999empirical, powel2020empiricalobservabilitygramianstochastic}---which generalizes the classical linear observability Gramian to nonlinear systems using only forward simulations---provides superior trajectory guidance over Cramér-Rao Lower Bound (CRLB) and Fisher Information Matrix (FIM) for single-sensor source seeking. We extend this to the 3D turbulent setting.

Two complementary Gramian computations are maintained. The \textit{observability Gramian} quantifies accumulated information about $\theta$ by perturbing the source estimate. At step $k$, the sensitivity vector $\boldsymbol{\phi}_k^\varepsilon \in \mathbb{R}^3$ is:
\begin{equation}
    \phi_{k,i}^\varepsilon = C(\mathbf{r}_k;\, \hat{\theta} + \varepsilon \mathbf{e}_i) - C(\mathbf{r}_k;\, \hat{\theta} - \varepsilon \mathbf{e}_i), \quad i = 1,2,3,
    \label{eq:emgr_phi}
\end{equation}
where $\mathbf{e}_i \in \mathbb{R}^3$ are the standard basis vectors and $\varepsilon$ is the perturbation magnitude. The local Gramian and its exponentially-weighted accumulation are \cite{powel2020empiricalobservabilitygramianstochastic}:
\begin{equation}
    \mathbf{W}_k^{\mathrm{local}} = \frac{1}{4\varepsilon^2}\, \boldsymbol{\phi}_k^\varepsilon\, (\boldsymbol{\phi}_k^\varepsilon)^T, \qquad \mathbf{W}_k = \gamma_w\, \mathbf{W}_{k-1} + \mathbf{W}_k^{\mathrm{local}},
    \label{eq:W_accum}
\end{equation}
with $\mathbf{W}_0 = \mathbf{0}_{3\times 3}$. Unlike the static 2D setting in \cite{giri2025optimal} where a cumulative sum sufficed, the drifting $\hat{\theta}$ in 3D turbulence necessitates the forgetting factor $\gamma_w < 1$. The reciprocal of the smallest singular value of the Gramian, $\varsigma_{\min}^{-1}$, known as the unobservability index \cite{krener2009measures}, quantifies the worst-case directional observability of $\theta$. For the RL observation vector, introduced in next section, the aggregate accumulated information is instead summarized by the inverse trace $\zeta_k^{-1} = 1/(\mathrm{tr}(\mathbf{W}_k) + \varepsilon_W)$, a computationally efficient scalar proxy where $\varepsilon_W$ is a small constant to avoid divide by zero error.

The \textit{planning Gramian} guides the deterministic baseline planner by perturbing the sensor position instead. For each candidate position $\mathbf{r}^{(j)}$, the planner evaluates:
\begin{equation}
    \psi_{j,i}^\varepsilon = C(\mathbf{r}^{(j)} + \varepsilon \mathbf{e}_i;\, \hat{\theta}) - C(\mathbf{r}^{(j)} - \varepsilon \mathbf{e}_i;\, \hat{\theta}), \quad i = 1,2,3,
    \label{eq:planner_phi}
\end{equation}
and selects the candidate minimizing $\varsigma_{\min}^{-1}$ of the resulting local Gramian.

\begin{remark}[Gramian Bias]
    Both Gramians are centered on $\hat{\theta}$, not the true $\theta$. When $\hat{\theta}$ is inaccurate, the sensitivity fields $\boldsymbol{\phi}_k^\varepsilon$ and $\boldsymbol{\psi}_j^\varepsilon$ are biased, creating an artificial attractive basin around $\hat{\theta}$. The planner navigates toward the estimate rather than the source, and $\mathcal{R}$ loses the spatial diversity required for the estimators to self-correct. This self-reinforcing degeneracy motivates the learned exploration in Section~\ref{Rl-framework}.
\end{remark}


\section{INFORMATION-GUIDED SAFE RL FRAMEWORK} \label{Rl-framework}

\begin{figure}[t]
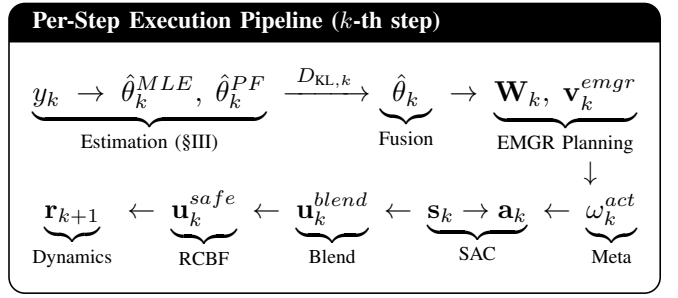

\begin{tcolorbox}[colback=white, colframe=black, boxrule=0.6pt, arc=2mm,
  left=2mm, right=2mm, top=2mm, bottom=2mm,
  title={\small\bfseries Per-Step Execution Pipeline ($k$-th step)},
  fonttitle=\small\bfseries, coltitle=white, colbacktitle=black]
\centering

\resizebox{\linewidth}{!}{%
$\displaystyle
\underbrace{y_k \;\to\; \hat{\theta}_k^{MLE},\;\hat{\theta}_k^{PF}}_{\text{Estimation~(\S\ref{estimation})}}
\;\xrightarrow{\;D_{\text{KL},k}\;}
\underbrace{\hat{\theta}_k}_{\text{Fusion}}
\;\to\;
\underbrace{\mathbf{W}_k,\;\mathbf{v}_k^{emgr}}_{\text{EMGR Planning}}
$%
}

\vspace{2pt}
\hfill$\downarrow$\hspace{4mm}
\vspace{2pt}

\resizebox{\linewidth}{!}{%
$\displaystyle
\underbrace{\mathbf{r}_{k+1}}_{\text{Dynamics}}
\;\leftarrow\;
\underbrace{\mathbf{u}_k^{safe}}_{\text{RCBF}}
\;\leftarrow\;
\underbrace{\mathbf{u}_k^{blend}}_{\text{Blend}}
\;\leftarrow\;
\underbrace{\mathbf{s}_k \to \mathbf{a}_k}_{\text{SAC}}
\;\leftarrow\;
\underbrace{\omega_k^{act}}_{\text{Meta}}
$%
}
\end{tcolorbox}
\vspace{-6pt}
\caption{Sequential execution pipeline. Sensor data feeds parallel estimators whose fused estimate drives both the classical EMGR planner and the learned SAC policy. The meta-supervisor allocates blending authority $\omega_k^{act}$; the safety layer enforces collision avoidance before actuation.}
\label{fig:pipeline}
\end{figure}

The Gramian bias identified in Remark~1 creates a self-reinforcing degeneracy: the planner navigates toward $\hat{\theta}$ rather than $\theta$, starving the estimators of spatial diversity. To break this cycle, we envelop the EMGR framework within a Reinforcement Learning architecture where a deterministic meta-supervisor dynamically blends the classical planner with a learned exploratory policy. Fig.~\ref{fig:pipeline} summarizes the per-step data flow before we develop each component. Throughout, we write $[x]_a^b = \max(a, \min(x, b))$ for scalar saturation, $[x]^{+} = \max(0, x)$, $\Pi_\Omega$ for projection onto the admissible domain, and $\omega$ (with various superscripts) for deterministic blending weights---distinguished from the particle filter weights $w_k^{(i)}$.

\subsection{Markov Decision Process Formulation}
We formulate source seeking as a discounted Markov Decision Process $(\mathcal{S}, \mathcal{A}, P, R, \gamma)$ with discount factor $\gamma \in (0,1)$.

\subsubsection{State Space ($\mathcal{S}$)}
The agent observes a $15$-dimensional vector $\mathbf{s}_k \in \mathbb{R}^{15}$:
\begin{equation}
    \mathbf{s}_k \!=\! \bigl[\,\tilde{\mathbf{r}}_k,\;\tilde{\mathbf{u}}_k^w,\;\tilde{c}_k,\;c_k^{det},\;\tilde{\mathbf{v}}_k^{emgr},\;\tilde{D}_k,\;\tilde{H}_k,\;\tilde{n}_k,\;\tilde{\zeta}_k^{-1}\,\bigr]^{\!\top}.
    \label{eq:state}
\end{equation}
The first eight features encode raw perception. $\tilde{\mathbf{r}}_k \in [0,1]^3$ is the position normalized by domain extents. $\tilde{\mathbf{u}}_k^w \in [0,1]^3$ is the local wind velocity scaled by $u_{max}^w$ and shifted to $[0,1]$. The scalar $\tilde{c}_k = [\ln(1 + c_k^{sig}/\kappa_c)\,/\,\ln(1 + c_{ceil}/\kappa_c)]_0^1$ is a log-compressed concentration, where $c_k^{sig} = [y_k - b]^{+}$ is the bias-corrected sensor signal, $\kappa_c$ is the noise scale, and $c_{ceil}$ the saturation ceiling. The binary flag $c_k^{det} = \mathbb{1}[c_k^{sig} > 3\sigma_e]$ indicates plume detection above the noise floor.

The EMGR planner velocity is injected as a directional hint, attenuated by estimator reliability:
\begin{equation}
    \tilde{\mathbf{v}}_k^{emgr} = \frac{\mathbf{v}_k^{emgr}}{v_{max}} \cdot \exp\!\bigl({-[D_{\text{KL},k} - \delta_h]^+}\bigr)
    \label{eq:emgr_hint}
\end{equation}
where $\delta_h$ is a deadzone threshold. As $D_{\text{KL}}$ grows, the exponential decays toward zero, signaling the agent that the planner is unreliable. The remaining features are: the normalized KL alarm $\tilde{D}_k = [D_{\text{KL},k}/D_{max}]_0^1$; the particle filter Shannon entropy $\tilde{H}_k = [H_k/H_{max}]_0^1$ with $H_k = -\sum_i w_k^{(i)} \ln w_k^{(i)}$; the void-step counter $\tilde{n}_k = [n_k^{void}/n_{max}]_0^1$ (steps since last detection); and $\tilde{\zeta}_k^{-1} = (1 + \zeta_k^{-1}/\zeta_{sc})^{-1}$, a sigmoidal compression of the unobservability index from $\mathrm{tr}(\mathbf{W}_k)$.

\subsubsection{Action Space ($\mathcal{A}$)}
Under first-order kinematics the control input to the sUAS is a velocity command. The Soft Actor-Critic policy outputs $\mathbf{a}_k \in [-1,1]^3$ via a squashed Gaussian, decoded to velocity as $\mathbf{v}_k^{RL} = v_{max} \cdot \mathbf{a}_k$. The agent outputs a full 3D velocity vector; blending with the classical planner is handled by the meta-supervisor below.

\subsubsection{Transition Dynamics ($P$)}
The sUAS position evolves as:
\begin{equation}
    \mathbf{r}_{k+1} = \Pi_\Omega\!\bigl(\mathbf{r}_k + \Delta t\,(\mathbf{u}_k^{safe} + \mathbf{d}_k)\bigr)
    \label{eq:dynamics}
\end{equation}
where $\mathbf{u}_k^{safe}$ is the safety-filtered velocity from Section~\ref{sec:rcbf}, $\mathbf{d}_k$ is a bounded disturbance with $\|\mathbf{d}_k\| \leq d_{max}$, and $\Pi_\Omega$ enforces domain bounds and altitude constraints.

\subsection{Deterministic Meta-Supervisor} \label{sec:meta}
The mission executes two phases. In \textit{Phase~0}, the sUAS runs a deterministic search---perimeter sweeps then wind-biased L\'{e}vy flights---until the plume is detected and the MLE activates. In \textit{Phase~1}, the RL agent augments the EMGR planner via deterministic blending governed by four handover signals.

When the particle filter is active, the MLE and particle filter estimates are fused according to their disagreement:
\begin{align}
    \hat{\theta}_k &= (1 - \omega_k^{fuse})\,\hat{\theta}_k^{MLE} + \omega_k^{fuse}\,\hat{\theta}_k^{PF}
    \label{eq:fusion} \\
    \omega_k^{fuse} &= \max\!\bigl(\sigma(\|\hat{\theta}_k^{MLE} \!-\! \hat{\theta}_k^{PF}\| - d_\sigma),\;\omega_k^{KL}\bigr)
    \label{eq:fusion_weight}
\end{align}
where $\sigma(\cdot)$ is the logistic sigmoid with midpoint $d_\sigma$, and
\begin{equation}
    \omega_k^{KL} = 1 - \exp\!\bigl({-[\kappa_{kl}\,D_{\text{KL},k} - b_{kl}]^+}\bigr)
    \label{eq:w_kl}
\end{equation}
is the KL-triggered exploration signal with sensitivity $\kappa_{kl}$ and bias $b_{kl}$, directly implementing the mismatch trigger from Section~\ref{estimation}. Three additional signals govern RL authority. The \textit{proximity} signal counters Gramian bias near $\hat{\theta}_k$, and the \textit{void} signal activates when the plume is lost:
\begin{align}
    \omega_k^{prox} &= 1 - \left[\frac{\|\mathbf{r}_k - \hat{\theta}_k\| - d_0}{d_1 - d_0}\right]_{\!0}^{1} \label{eq:w_prox} \\[2pt]
    \omega_k^{void} &= \left[\frac{n_k^{void} - n_0}{n_1 - n_0}\right]_{\!0}^{1} \label{eq:w_void}
\end{align}
where $d_0, d_1$ are proximity onset and saturation distances, and $n_0, n_1$ are void-step onset and saturation counts. The composite blending weight and resulting velocity command are:
\begin{align}
    \omega_k^{act} &= \max\!\bigl(\omega_{fl},\;\omega_k^{KL},\;c_p\,\omega_k^{prox},\;\omega_k^{void}\bigr) \label{eq:w_action} \\
    \mathbf{u}_k^{blend} &= (1 - \omega_k^{act})\,\mathbf{v}_k^{emgr} + \omega_k^{act}\,\mathbf{v}_k^{RL} \label{eq:blend}
\end{align}
where $\omega_{fl}$ is a minimum exploration floor and $c_p$ scales the proximity term. A stagnation override sets $\omega_k^{act} = 1$ for $n_{esc}$ steps whenever displacement falls below $\delta_{stag}$ for $n_{stag}$ consecutive steps.

\subsection{Soft Actor-Critic Policy} \label{sec:sac}
The policy $\pi_\phi$ maximizes the entropy-regularized return~\cite{haarnoja2018soft}:
\begin{equation}
    J(\pi) = \sum_{k=0}^{\infty} \gamma^k\,\mathbb{E}_\pi\!\bigl[R_k + \eta\,\mathcal{H}\bigl(\pi(\cdot|\mathbf{s}_k)\bigr)\bigr]
    \label{eq:sac_objective}
\end{equation}
where $\eta > 0$ is the entropy temperature. The entropy bonus prevents the policy from collapsing to a single trajectory mode. The actor parameterizes a squashed Gaussian:
\begin{equation}
    \mathbf{a}_k = \tanh\!\bigl(\boldsymbol{\mu}_\phi(\mathbf{s}_k) + \boldsymbol{\sigma}_\phi(\mathbf{s}_k) \odot \boldsymbol{\epsilon}\bigr),\;\;\boldsymbol{\epsilon} \sim \mathcal{N}(\mathbf{0},\mathbf{I}).
    \label{eq:actor}
\end{equation}
Twin Q-networks $Q_{\psi_1}, Q_{\psi_2}$ mitigate overestimation via the soft Bellman target:
\begin{equation}
    G_k \!=\! R_k + \gamma\!\left[\min_{j} Q_{\bar{\psi}_j}\!(\mathbf{s}_{k+1}, \tilde{\mathbf{a}}) - \eta \ln \pi_\phi(\tilde{\mathbf{a}}|\mathbf{s}_{k+1})\right]
    \label{eq:q_target}
\end{equation}
for non-terminal transitions, where $\tilde{\mathbf{a}} \sim \pi_\phi(\cdot|\mathbf{s}_{k+1})$ and $\bar{\psi}$ denotes Polyak-averaged target parameters with rate $\rho$. The critic minimizes the Huber loss for robustness to large terminal rewards. The temperature $\eta$ is auto-tuned toward target entropy $\bar{\mathcal{H}} = -\dim(\mathcal{A})/2$ with a hard floor $\eta \geq \eta_{min}$ to prevent exploration collapse.

\subsection{Robust Control Barrier Functions} \label{sec:rcbf}
Let $h_i(\mathbf{r})$ be the signed distance to the $i$-th obstacle ($h_i > 0$ exterior), inflated by a keepout margin $\delta_s$ as $h_i^{eff} = h_i - \delta_s$. The safe velocity command is obtained via a minimal-correction Quadratic Program~\cite{ames2019control, cheng2019end}:
\begin{equation}
\begin{aligned}
    \mathbf{u}_*^{S} &= \arg\min_{\mathbf{u}^{S},\,\varepsilon_s} \;\;\tfrac{1}{2}\|\mathbf{u}^{S}\|^2 + M\,\varepsilon_s \\
    \text{s.t.}\;\; &\nabla h_i^T(\mathbf{u}_k^{blend} \!+\! \mathbf{u}^{S}) \geq \\[-2pt]
    &\quad {-}\gamma_b\,h_i^{eff} + \|\nabla h_i\|\,d_{max} - \varepsilon_s,\\
    &\varepsilon_s \geq 0,\quad \forall\,i
\end{aligned}
\label{eq:rcbf_qp}
\end{equation}
where $\gamma_b > 0$ is the class-$\mathcal{K}$ decay rate, $M \gg 1$ penalizes constraint relaxation, and the $\|\nabla h_i\|\,d_{max}$ term robustifies against worst-case disturbances~\cite{xu2015robustness}. The applied velocity is $\mathbf{u}_k^{safe} = \mathbf{u}_k^{blend} + \mathbf{u}_*^{S}$, solved via warm-started OSQP~\cite{stellato2020osqp}. The RL agent trains on the executed safe action $\mathbf{u}_k^{safe}/v_{max}$ rather than its proposed $\mathbf{a}_k$, teaching it to anticipate safety corrections.

\subsection{Curriculum Training Strategy} \label{sec:curriculum}
Training directly in the full setting (large domain, sparse plume, mobile source) yields prohibitively slow convergence due to reward sparsity. To mitigate this combinatorial difficulty, we adopt a progressive three-stage curriculum, parameterizing the source spawn region $\Omega_{spawn} \subseteq \Omega$ and velocity bound $\|\dot{\theta}\| \leq v_{drift}$:

\begin{enumerate}
    \item \textbf{Stage 1} ($\Omega_{spawn} \subset \Omega,\; v_{drift} \!=\! 0$)\textbf{:} The static source is restricted to a central sub-domain. High plume encounter rates yield dense rewards, establishing core tracking primitives: EMGR gradient ascent, KL-triggered handovers, and CBF anticipation.
    
    \item \textbf{Stage 2} ($\Omega_{spawn} = \Omega,\; v_{drift} \!=\! 0$)\textbf{:} The agent fine-tunes the Stage 1 policy across the full domain, resolving prolonged search phases and attenuated initial concentration signals.
    
    \item \textbf{Stage 3} ($\Omega_{spawn} = \Omega,\; \|\dot{\theta}\| \leq v_{drift}$)\textbf{:} Stochastic source drift is introduced. This full setting trains robustness to non-stationary targets, demanding continuous re-acquisition and active void-step recovery.
\end{enumerate}

Letting $\pi_i$ denote the converged policy of stage $i$, each stage initializes via $\pi_i \leftarrow \pi_{i-1}$ and trains until the success rate plateaus, systematically circumventing the full task's combinatorial complexity.

\subsection{Reward Design and Convergence}
The particle filter posterior shift between steps is measured via the marginal 1-Wasserstein distance:
\begin{equation}
    R_k^{W_1} = \Bigl[\mu_w\bigl[\Delta W_{1,k} - \delta_j\bigr]^{+}\Bigr]_{\!0}^{R_{max}}
    \label{eq:r_emd}
\end{equation}
where $\Delta W_{1,k} = \sum_{d=1}^{3} W_1(\hat{P}_k^{(d)},\,\hat{P}_{k-1}^{(d)})$ is the total coordinate-wise Wasserstein shift, $\delta_j$ subtracts resampling jitter, and $\mu_w$ is a scaling gain. A proximity shaping term
\begin{equation}
    R_k^{prog} = \bigl[\mu_p\bigl(\|\mathbf{r}_{k-1} - \theta\| - \|\mathbf{r}_k - \theta\|\bigr)\bigr]_{-1}^{\,1}
    \label{eq:r_prog}
\end{equation}
uses the true source position $\theta$ as \textit{privileged information} to accelerate credit assignment in simulation. This potential-based reward shaping~\cite{ng1999policy} provides dense feedback while theoretically guaranteeing the optimal policy remains unchanged. Crucially, $\theta$ strictly informs the scalar reward and never enters the agent's observable state $\mathbf{s}_k$~\eqref{eq:state}. During deployment evaluation, $\theta$ remains unknown and unaccessed, ensuring real-world applicability.

A safety penalty $R_k^{safe} = -\lambda_s\|\mathbf{u}_*^{S}\|$ discourages reliance on the barrier layer. The composite step reward, including time cost $\xi < 0$, is:
\begin{equation}
    R_k = \bigl[R_k^{W_1} + R_k^{prog} + \xi + R_k^{safe}\bigr]_{\!-R_{cl}}^{\,R_{cl}}.
    \label{eq:r_step}
\end{equation}
Terminal rewards $R^{term} \!\in\! \{R_+,\,{-}R_o,\,{-}R_c\}$ for success ($\|\mathbf{r}_k \!-\! \theta\| < d_{conv}$), timeout, and collision satisfy $R_+ \gg R_o > R_c > R_{cl}$. Because the RCBF layer~\eqref{eq:rcbf_qp} structurally enforces collision avoidance independently of the reward, the collision penalty acts merely as a soft shaping signal. 

During training, episode termination is triggered by an oracle proximity check $\|\mathbf{r}_k - \theta\| < d_{conv}$. Like $R_k^{prog}$, this leverages privileged simulation data to accelerate learning and serves solely as a training surrogate. For onboard deployment, we propose a stopping rule leveraging a non-decaying cumulative Gramian $\mathbf{W}_k^c = \mathbf{W}_{k-1}^c + \mathbf{W}_k^{local}$. Based on physical UGV trials~\cite{giri2025lowcost}, $\varsigma_{max}(\mathbf{W}_k^c)$ rises sharply near the source, suggesting termination when $\varsigma_{max}(\mathbf{W}_k^c) \geq \varsigma_{th} \land c_k^{det} > 0$. While calibrating this sUAS criterion remains ongoing, Section~\ref{sec:results} uniformly applies the oracle threshold $d_{conv}$ for evaluation. 

Finally, timeouts are \emph{not} treated as true terminal states in the replay buffer; Bellman backups bootstrap from successor states to distinguish environmental termination from artificial time limits~\cite{pardo2018time}. Algorithm~\ref{alg:meta_sac_train} details the full execution loop.

\begin{algorithm}[tb]
\small
\caption{Information-Guided Residual SAC Training}\label{alg:meta_sac_train}
\KwIn{Search budget $T_s$, horizon $T_{max}$, convergence radius $d_{conv}$}
\KwOut{Trained policy $\pi_\phi$, critics $Q_{\psi_{1,2}}$}
Initialize policy $\pi_\phi$, critics $Q_{\psi_{1,2}}$, targets $Q_{\bar\psi_{1,2}}$, replay buffer $\mathcal{B} \leftarrow \emptyset$, temperature $\eta$\;
\For{$e = 1, \dots, N_{ep}$}{
    Sample initial position $\mathbf{r}_0 \sim \rho_0$ and source $\theta \sim \mathcal{U}(\Omega)$\;
    Set $\mathbf{W}_0^c \leftarrow \mathbf{0}_{3\times 3}$\;
    \tcp{Phase 0: Deterministic Search}
    \For{$k = 0, \dots, T_s - 1$}{
        Acquire measurement $y_k$ via~\eqref{eq:observation_model}\;
        \lIf{$c_k^{sig} > 3\sigma_e$}{update MLE~\eqref{eq:proj_mle} and PF~\eqref{eq:pf_update}}
        Compute velocity from perimeter sweep or L\'{e}vy flight\;
        Apply RCBF safety filter~\eqref{eq:rcbf_qp} and step dynamics~\eqref{eq:dynamics}\;
        \lIf{plume detected $\land$ estimate availabe}{\textbf{enter Phase\,1}}
    }
    \tcp{Phase 1: RL-Augmented Tracking}
    \For{$k = 0, \dots, T_{max} - 1$}{
        Acquire measurement $y_k$ and update MLE, PF\;
        Evaluate mismatch $D_{\text{KL},k}$~\eqref{eq:kl_mismatch} and fuse $\hat{\theta}_k$~\eqref{eq:fusion}\;
        Accumulate Gramians $\mathbf{W}_k$, $\mathbf{W}_k^c$~\eqref{eq:W_accum}\;
        Compute planner velocity $\mathbf{v}_k^{emgr}$~\eqref{eq:planner_phi}\;
        Evaluate blending weight $\omega_k^{act}$~\eqref{eq:w_action}\;
        Construct state $\mathbf{s}_k$~\eqref{eq:state} and sample $\mathbf{a}_k \sim \pi_\phi(\cdot|\mathbf{s}_k)$\;
        Set $\mathbf{v}_k^{RL} = v_{max}\,\mathbf{a}_k$ and blend~\eqref{eq:blend}\;
        Apply RCBF~\eqref{eq:rcbf_qp} and step dynamics~\eqref{eq:dynamics}\;
        Evaluate reward $R_k$~\eqref{eq:r_step}\;
        Store $(\mathbf{s}_k,\,\mathbf{u}_k^{safe}/v_{max},\,R_k,\,\mathbf{s}_{k+1})$ in $\mathcal{B}$\;
        \lIf{$|\mathcal{B}| \geq B$}{update $\phi$, $\psi$, $\eta$ via SAC}
        \lIf{$\|\mathbf{r}_k - \theta\| < d_{conv}$ or collision}{apply $R^{term}$ and \textbf{break}}
    }
}
\end{algorithm}

\section{SIMULATION EXPERIMENTS AND RESULTS} \label{sec:results}

\subsection{Experimental Setup}

The domain is $\Omega = 256 \times 128 \times 128$\,m$^3$ under Pasquill--Gifford stability class D\cite{NOAA_READY}, with 3 to 5 randomly placed spherical obstacles per episode. The sUAS flies between $z \in [4, 60]$\,m at maximum velocity, $v_{max} = 8$\,m/s with $\Delta t = 0.1$\,s. Each curriculum stage trains for 500 episodes with a search budget of $T_s = 1000$ steps and a tracking horizon of $T_{max} = 2000$ steps. Stage~1 restricts the static source to a central $80 \times 40$\,m region, roughly 10\% of the domain. Stage~2 expands the source domain to $226 \times 98$\,m for random spawn on each episode while keeping the source static. Stage~3 retains the random spawn on same domain as Stage~2 and introduces stochastic source drift via an Ornstein--Uhlenbeck process\cite{uhlenbeck1930theory} with $\theta_{OU} = 0.03$ and $\sigma_{OU} = 0.4$\,m/$\sqrt{\text{s}}$. All five configurations are evaluated over 100 episodes with identical convergence criteria $\|\mathbf{r}_k - \theta\| < 5$\,m.

\subsection{Results and Discussions}

\begin{table}[H]
\centering
\caption{Localization Performance (100 episodes, $d_{conv} = 5$\,m)}
\label{tab:performance}
\setlength{\tabcolsep}{3pt}
\begin{tabular}{lccc}
\hline
\textbf{Method} & \textbf{Success} & \textbf{Med.\ Dist.\ [5th, 95th]} & \textbf{Med.\ Steps}\\
\hline
MLE+EMGR          & 30.4\% & 24.3\,[4.7, 135.8]\,m & 1173 \\
MLE+EMGR+PF+KL    & 32.1\% & 25.0\,[4.8, 159.3]\,m & 2000 \\
\hline
Stage 1 (small)    & 98.8\% & 4.9\,[4.7, 5.0]\,m    & 256  \\
Stage 2 (full)     & 79.3\% & 4.9\,[4.7, 43.8]\,m   & 310  \\
Stage 3 (mobile)   & 78.6\% & 4.8\,[4.5, 59.2]\,m   & 276  \\
\hline
\end{tabular}
\end{table}

Table~\ref{tab:performance} confirms the central thesis. Both baselines stall around 30\% success with median distances above 24\,m, demonstrating that the Gramian bias of Remark~1 cannot be resolved by belief diversity alone---the PF+KL ablation exhausts the full 2000-step horizon without meaningful improvement. The RL agents break this degeneracy decisively. Stage~1 reaches 98.8\% success in a median of 256 steps, though this near-perfect rate partly reflects overfitting to the restricted spawn region. Expanding to the larger domain in Stage~2 drops success to 79.3\%, quantifying the generalization gap: the core tracking skill transfers, but the agent must now handle distant sources and weaker initial signals across a seven times larger area. Stage~3 maintains 78.6\% despite mobile sources with the fastest median convergence at 276 steps, confirming that the curriculum transfers tracking primitives---EMGR following, KL-triggered handovers, CBF anticipation---rather than memorized trajectories.

\begin{figure}[t]
    \centering
    \includegraphics[width=0.8\linewidth]{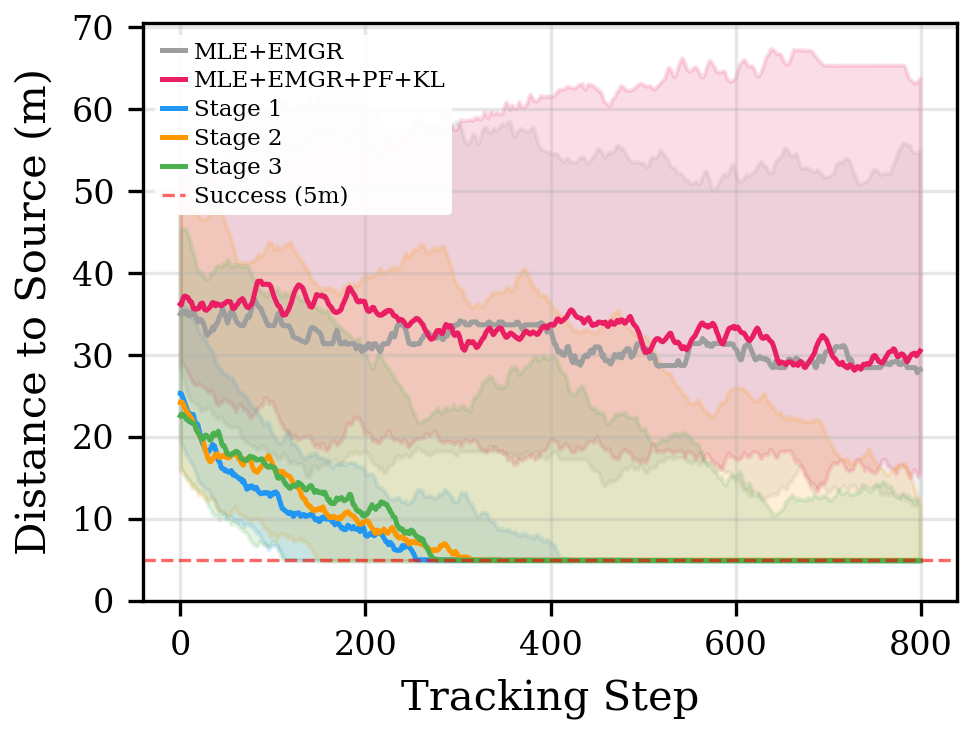}
    \caption{Median distance to source with interquartile bands. RL agents converge below $d_{conv} = 5 m$ within 300 steps; baselines stagnate due to Gramian bias.}
    \label{fig:convergence}
\end{figure}

\begin{figure}[t]
    \centering
    \includegraphics[width=0.8\linewidth]{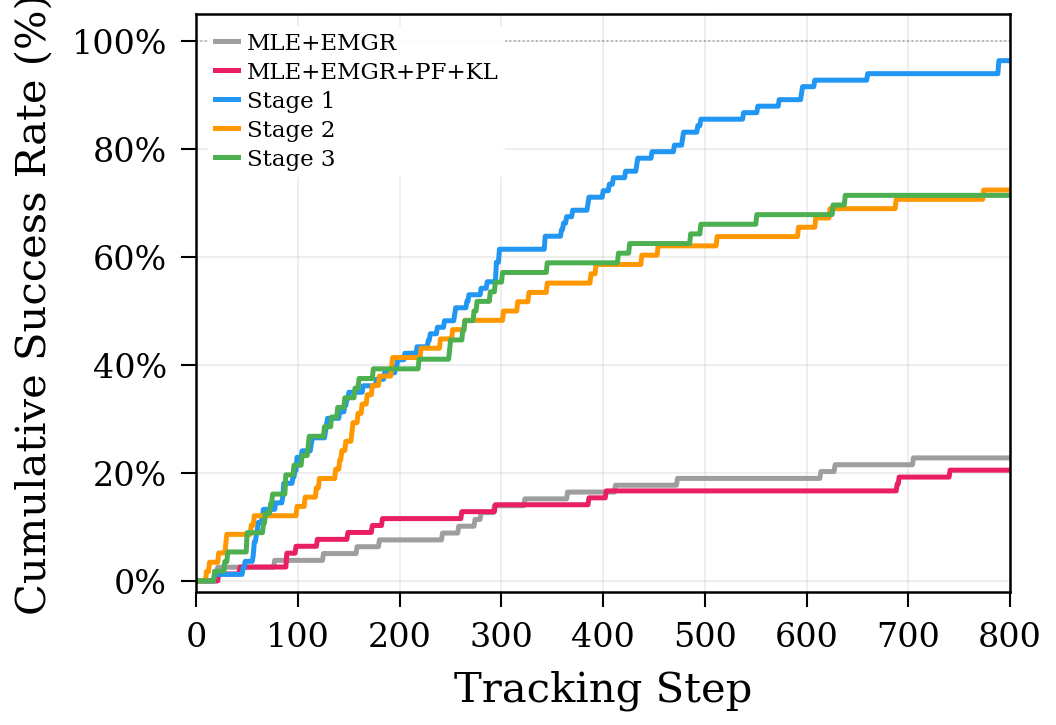}
    \caption{Cumulative success rate over tracking steps.}
    \label{fig:cum_success}
\end{figure}

Fig.~\ref{fig:convergence} shows that all RL stages drive the median distance below $d_{conv}$ within 300 steps while baselines plateau above 20\,m. Fig.~\ref{fig:cum_success} confirms the RL agents reach their terminal success rates three to four times faster than baselines.

Table~\ref{tab:safety} reports zero barrier violations across all episodes where plume was detected. The RL agents trigger the RCBF far more frequently---84 to 89\% of steps versus 7 to 13\% for baselines---due to aggressive exploration near obstacles, yet the minimum barrier margin never drops below 2.54\,m. This confirms that the safety guarantee from~\eqref{eq:rcbf_qp} is structural and independent of the learned policy. Simulation for a random episode after plume is detected (only Phase 1) for each of these methods is uploaded at: \url{https://youtu.be/etPAydxxQGs}. Note that perimeter sweep and Lévy flights are omitted from this video due to their significantly longer runtimes.

\begin{table}[H]
\centering
\caption{RCBF Safety Statistics}
\label{tab:safety}
\setlength{\tabcolsep}{3pt}
\begin{tabular}{lcccc}
\hline
\textbf{Method} & \textbf{Active} & \textbf{Max Corr.} & \textbf{Min $h(\mathbf{r})$} & \textbf{Violations}\\
\hline
MLE+EMGR       & 12.9\% & 4.87 & 2.54\,m & 0 \\
MLE+EMGR+PF+KL & 7.2\%  & 4.84 & 2.54\,m & 0 \\
\hline
Stage 1         & 88.9\% & 6.91 & 3.04\,m & 0 \\
Stage 2         & 84.3\% & 5.77 & 3.04\,m & 0 \\
Stage 3         & 88.4\% & 5.59 & 3.04\,m & 0 \\
\hline
\end{tabular}
\end{table}

One limitation of the convergence criterion is that it relies on oracle distance, which is unavailable in deployment. Calibrating a deployable Gramian-based stopping rule for sUAS is ongoing work. The 20\% failure rate in Stages~2 and~3 stems primarily from boundary sources and occluded plumes that prevent reliable detection within the search budget. 


\section{CONCLUSION} \label{conclusion}

This study presented an Information-Guided Safe Reinforcement Learning framework for autonomous trace gas localization. Because source estimation in turbulence is inherently ill-posed, classical information-seeking strategies often stall in local degeneracies. To overcome this, we enveloped a classical EMGR planner within a curriculum-trained Meta-SAC architecture. Governed by real-time KL divergence monitoring, a meta-supervisor dynamically blends deterministic planning with learned exploration, actively driving the sUAS into high-information zones to regularize the estimates. Validated in a GPU-accelerated 3D simulation with a strict RCBF safety filter, this framework provides a rigorous foundation for robust deployment. Future work includes sim-to-real hardware transfer and multi-source, multi-agent scenarios.


\section*{REPRODUCIBILITY}
The code for this environment is provided at GitHub: \url{https://github.com/sachingirime/Info-guided-safe-RL.git} and the trained models for all 3 RL stages are uploaded to Kaggle: \url{https://www.kaggle.com/models/sachingiri348/info-guided-safe-rl}


\bibliographystyle{IEEEtran}
\bibliography{mybib}
\end{document}